\documentclass{article}

\usepackage{spconf,amsmath,graphicx,hyperref}

\usepackage{enumitem}
\usepackage{tabularray}

\usepackage{multirow}
\usepackage{makecell}
\usepackage{graphicx}
\usepackage{subcaption}

\usepackage[normalem]{ulem}

\usepackage{booktabs}
\usepackage{caption}
\usepackage{array}
\usepackage{svg}

\title{COCO-Inpaint: A Benchmark for Detecting and Localizing Inpainting-Based Image Manipulations}
\name{}
\address{}

\name{\shortstack{Haozhen Yan$^{1}$,
      Yan Hong$^{2}$,
      Jiahui Zhan$^{1}$,
      Suning Lang$^{1}$,
      Yikun Ji$^{1}$,\\
      Yujie Gao$^{1}$,
      Huijia Zhu$^{2}$,
      Jun Lan$^{2}$\sthanks{Corresponding authors.},
      Jianfu Zhang$^{1}$\footnotemark[1]}}
\address{$^{1}$ Shanghai Jiao Tong University, $^{2}$ Ant Group
\\ \small orion810@sjtu.edu.cn, lanjun\_yelan@163.com, c.sis@sjtu.edu.cn
}

\begin{document}
%
\maketitle
\begin{abstract}
Recent advances in image manipulation have enabled highly photorealistic content generation, but also lowered the barrier to arbitrary editing, raising concerns about multimedia authenticity and security. Existing Image Manipulation Detection and Localization (IMDL) methods mainly target splicing or copy-move forgeries, while benchmarks for inpainting-based manipulations remain limited. To bridge this gap, we present COCO-Inpaint, a comprehensive benchmark specifically designed for inpainting detection and localization, with three key contributions: 1) High-quality inpainting samples generated by six state-of-the-art inpainting models, 2) Diverse generation scenarios enabled by four mask generation strategies with optional text guidance, and 3) Large-scale coverage of 238,302 inpainted images with rich semantic diversity. Our benchmark is constructed to highlight intrinsic inconsistencies between inpainted and authentic regions, rather than superficial semantic artifacts such as object shapes. We further establish a rigorous evaluation protocol with three standard metrics to benchmark existing IMDL methods and reveal current trends and challenges.
\end{abstract}
\section{Introduction}
\label{sec:intro}

With the rapid development of generative methods~\cite{BrushNet_2024, PowerPaint_2024}, image generation has achieved remarkable progress in recent years.
Image inpainting, as a fundamental task in image generation, aims to fill in the missing regions of an image with plausible content.
However, inpainting methods present a double-edged sword: while their operational simplicity and impressive results empower users with intuitive editing capabilities, these very advantages also raise significant concerns regarding digital integrity and information security when misused.
This urgently demands effective methods to detect and localize inpainting manipulations.

\begin{figure}[t]
    \centering
    \includegraphics[width=\linewidth]{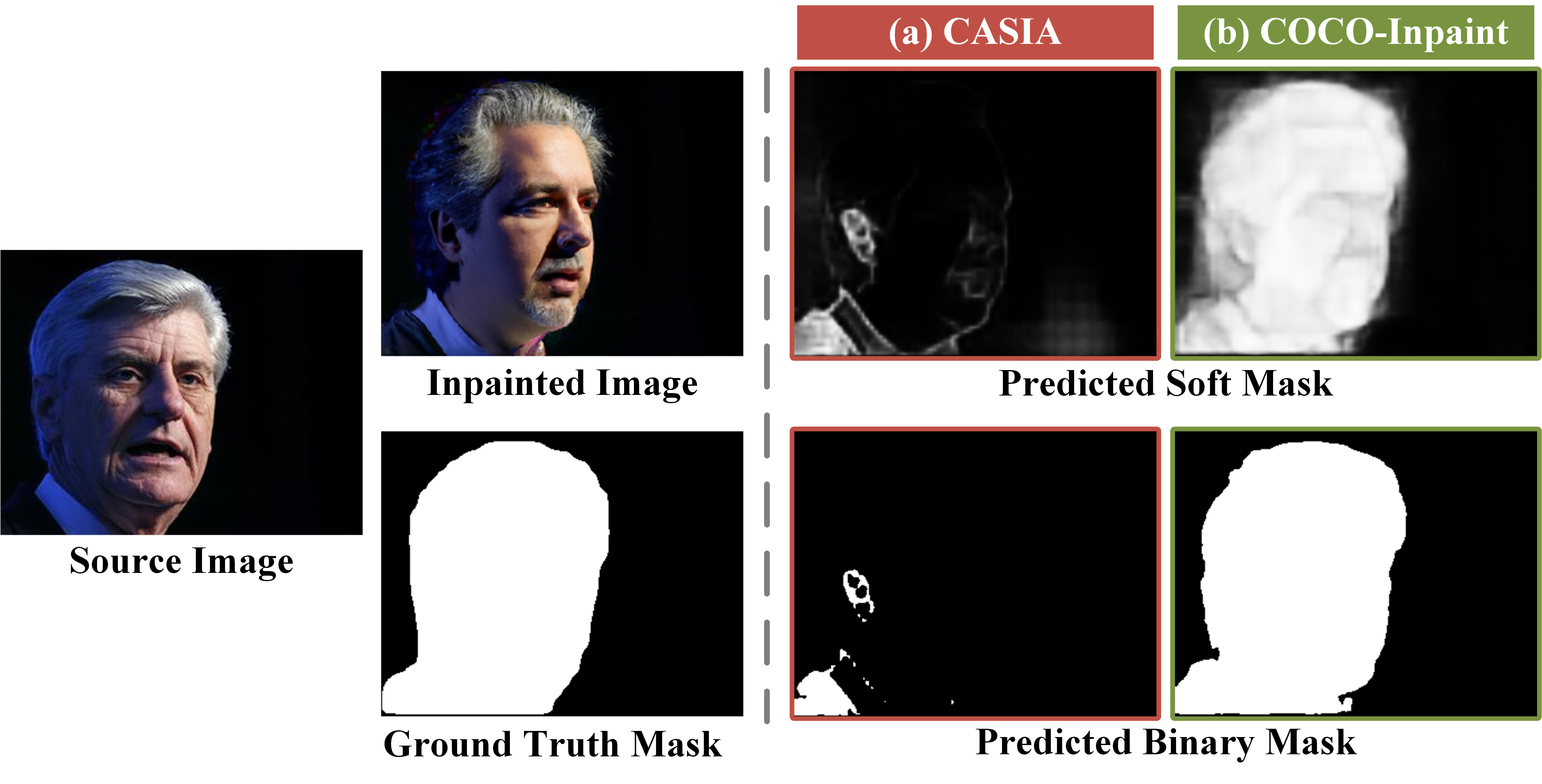}
    \vspace{-16pt}
    \caption{
    Comparison of IMDL model performance based on different training datasets. We select a source image from AutoSplice~\cite{AutoSplice_2023} and utilize DALL-E2~\cite{DALLE2_2022} to repaint it. (a) presents the prediction of localization using IML-ViT~\cite{IML-ViT_2023} trained on CASIA~\cite{CASIA_2013} while (b) is trained on COCO-Inpaint.
    Both the soft mask and binary mask are provided. Training on COCO-Inpaint substantially enhances the model’s detection sensitivity and segmentation accuracy.
    }
    \vspace{-16pt}
    \label{fig:dataset_example}
\end{figure}

Existing Image Manipulation Detection and Localization (IMDL) datasets~\cite{CASIA_2013,TampCOCO_2022,HiFi-IFDL_2023, IMD_2020} mainly generate manipulated images by normal {\em splicing} and {\em copy-move} operations (\textit{i.e.}, insert objects into other images, or duplicate regions within the same image) from segmentation datasets~\cite{COCO_2014}.
These two operations fail to capture the distinct characteristics of different images, limiting the generalizability of existing IMDL methods, as shown in Fig.~\ref{fig:dataset_example}.
Other existing studies have also focused on inpainting manipulation. 
AutoSplice~\cite{AutoSplice_2023} generates 3621 inpainting images from the Visual News dataset using an object segmentation pipeline and a generative model~\cite{DALLE2_2022}.
COCOGlide~\cite{COCOGlide_2023} includes 512 inpainting images from MS-COCO~\cite{COCO_2014} validation set by GLIDE~\cite{GLIDE_2021}.
GIM~\cite{GIM_2024} introduced the IMDL Benchmark, where manipulated images were generated using diffusion-based generative models.
However, these datasets remain limited in scale, quality, model diversity, and systematic analysis.
To overcome these limitations, we introduce \textbf{COCO-Inpaint}, a large-scale, multi-dimensional, high-quality inpainting IMDL benchmark.
Built on MS-COCO~\cite{COCO_2014}, COCO-Inpaint incorporates multiple representative inpainting models, four mask types (segmentation, bounding box, random polygon, random box), and two guidance modes (context-aware and text-guided).
The final dataset consists of $\sim\!238k$ inpainted images and $\sim\!108k$ authentic images, supporting fine-grained evaluation across all model-mask-guidance combinations.
As shown in Fig.~\ref{fig:dataset_example}, training IMDL models on COCO-Inpaint substantially improves localization of manipulated regions, whereas training on splicing- or copy-move-based datasets such as CASIA~\cite{CASIA_2013} yields limited performance.
We further conduct cross-validation experiments across different levels, leading to the following key observations:\looseness=-1

\begin{itemize}[leftmargin=*, noitemsep]

  \item  \textit{Model architecture:} Vision Transformer-based models consistently outperform CNN-based models in adaptability to IMDL tasks across diverse experimental conditions.

  \item  \textit{Cross-model generalization:} IMDL models perform well in-distribution but generalize poorly to unseen inpainting models, while multi-model training improves adaptability.

  \item  \textit{Cross-mask evaluation:} The shape of the mask significantly affects generalization. Models trained on randomly generated masks (\textit{e.g.}, random boxes or polygons) exhibit higher robustness than those trained on structured masks.
  
 \item \textit{Cross-mask-ratio evaluation:} Models trained with mask ratios between 0.4 and 0.6 achieve better generalization. A balanced proportion of inpainted and authentic regions encourages learning more discriminative features.

  \item \textit{Cross-prompt evaluation:} IMDL models trained on text-guided inpainted images achieve stronger generalization.

\end{itemize}

\begin{figure}[t]
    \centering
    \includegraphics[width=\linewidth]{fig/COCOInpaint_dataset_construct.pdf}
    \vspace{-8pt}
    \caption{COCO-Inpaint construction pipeline. Real images are converted into conditional images with 4 mask types. Optional COCO captions for text guidance. Six inpainting models then generate $6\times4\times2=48$ manipulated images.}
    \vspace{-8pt}
    \label{fig:dataset_construct}
\end{figure}

\section{COCO-Inpaint}
\label{sec:benchmark}
To build a high-quality IMDL benchmark, we consider image quality, mask/context diversity, and generalizability, and introduce COCO-Inpaint, a large-scale benchmark for inpainting-based manipulation analysis (Fig.~\ref{fig:dataset_construct}). \looseness=-1

\subsection{Inpainting Models}
To build a high-quality IMDL benchmark, we aim to cover diverse backgrounds with corresponding bounding boxes and segmentation masks. We adopt MS-COCO~\cite{COCO_2014} as the source dataset due to its rich semantic content and well-annotated object information.
Considering both performance and community adoption, six representative inpainting models are selected for manipulating real images, including three inpainting backbones: SD1.5-Inpainting~\cite{SD_2022}, SDXL-Inpainting~\cite{SDXL_2023} and Flux.1-Fill-dev~\cite{Flux_2024}, and two refined inpainting models: BrushNet~\cite{BrushNet_2024} and PowerPaint~\cite{PowerPaint_2024}.
Stable-Diffusion-3.5-large~\cite{SD_2022}, as a non-inpaint-specific model, is also selected for its powerful inpainting ability.

\begin{figure}[t]
    \centering

    \begin{subfigure}[t]{0.48\linewidth}
        \centering
        \includegraphics[width=\linewidth]{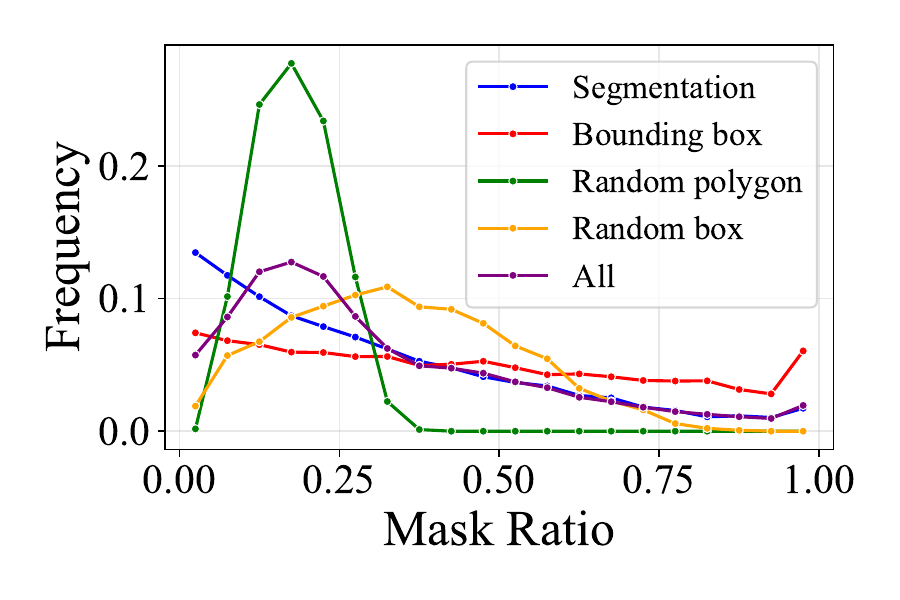}
        \vspace{-20pt}
        \caption{}
        \label{fig:mask_ratio}
    \end{subfigure}
    \hfill%
    \begin{subfigure}[t]{0.51\linewidth}
        \centering
        \includegraphics[width=\linewidth]{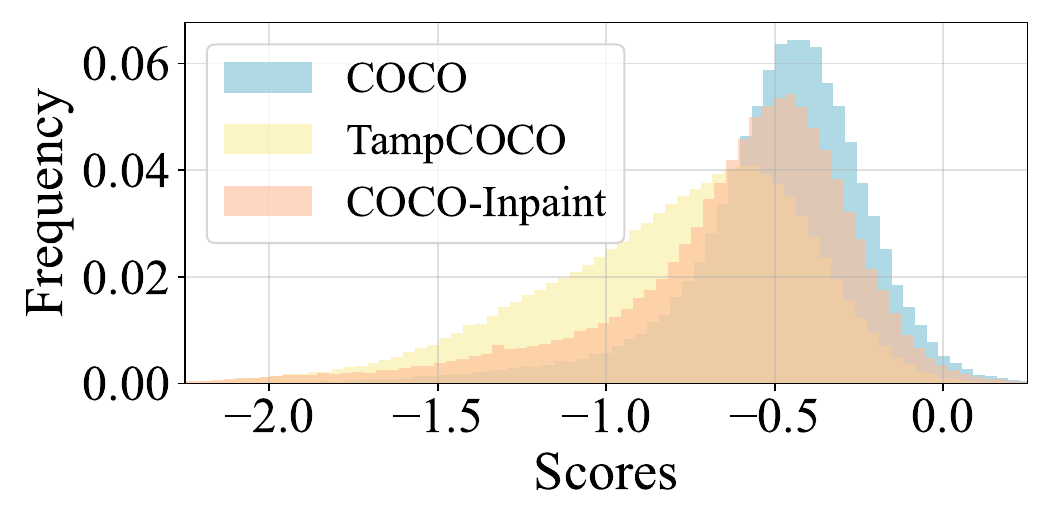}
        \vspace{-20pt}
        \caption{}
        \label{fig:image_reward}
    \end{subfigure}%

    \par\vspace{4pt}

    \begin{subfigure}[t]{0.51\linewidth}
        \centering
        \resizebox{\linewidth}{!}{
        \begin{tblr}{
          column{2} = {c},
          column{3} = {c},
          colsep=2pt,
          hline{1,6} = {-}{0.08em},
          hline{2} = {-}{lr},
        }
        Subset         & Text-guided & Context-aware \\
        Segmentation   & 5.06         & 17.15         \\
        Bounding Box   & 7.04         & 31.86         \\
        Random Polygon & 2.91         & 3.12          \\ 
        Random Box     & 3.91         & 9.20          
        \end{tblr}
        }
        \vspace{-4pt}
        \caption{}
        \label{tab:fid} 
    \end{subfigure}

    \vspace{-8pt}
    \caption{(a): Mask ratio distribution of COCO-Inpaint.
    (b): ImageReward distribution of COCO, TampCOCO and COCO-Inpaint. 
    (c): FID~($\downarrow$) under prompt guidance.}
    \vspace{-8pt}
    \label{fig:combined}
\end{figure}

\begin{table}[t]
\centering
\caption{Summary of IMDL Datasets. HiFi-IFDL mostly contains fully synthesized images; only manipulated ones are counted here. "\#Variants" denotes the number of manipulated images per authentic image.}
\vspace{-8pt}
\resizebox{0.95\linewidth}{!}{
\small
\begin{tblr}{
  column{even} = {c},
  column{3} = {c},
  column{5} = {c},
  hline{1,10} = {-}{0.08em},
  hline{2} = {1}{lr},
  hline{2} = {2-6}{},
}
Dataset      & \#Real & \#Fake & Generator & \#Model & \#Variants \\
CASIA~\cite{CASIA_2013}        & 7.2K   & 5.1K    & Trad.     & -       & 0-8           \\
IMD2020~\cite{IMD_2020}      & 35K    & 35K     & Trad.     & -       & 1           \\
TampCOCO~\cite{TampCOCO_2022}     & -      & 800K    & Trad.     & -       & 8         \\
HiFi-IFDL~\cite{HiFi-IFDL_2023}    & -      & 376K    & GAN       & 2       & 1           \\
AutoSplice~\cite{AutoSplice_2023}   & 2.3K   & 3.6K    & Diff.     & 1       & 1-3      \\
COCOGlide~\cite{COCOGlide_2023}    & -      & 512     & Diff.     & 1       & 1           \\
GIM~\cite{GIM_2024}          & 1.1M   & 1.1M    & Diff.     & 3       & 1           \\
COCO-Inpaint & 108K   & 238K    & Diff.     & 6       & 1 or 48     
\end{tblr}
}

\vspace{-16pt}
\label{tab:dataset}
\end{table}

\subsection{Masks and Prompts}
Masks are essential in inpainting as they define regions to be filled and encode different guidance signals. Segmentation masks enable object replacement or removal with prompts, while random masks promote background-consistent filling. Prior IMDL datasets~\cite{CASIA_2013,TampCOCO_2022} rely mainly on segmentation masks, whereas practical inpainting involves more diverse types. To reflect this diversity, we introduce four mask generation strategies: 
segmentation masks extracted from MS-COCO objects (merging multiple instances when needed), bounding boxes derived from object masks (merged into unified regions), and randomly generated polygons or boxes with 3–8 shapes each. 
Existing datasets such as CASIAv2 and Defacto contain small manipulated regions, averaging only 7.6\% and 1.7\% of image area, which limits evaluation under complex conditions. To address this, we merge multiple Random Polygon or Random Box masks into composite ones, enabling larger and more diverse mask ratios (Fig.~\ref{fig:mask_ratio}).

Recent inpainting models leverage prompts to enhance visual quality and semantic consistency. To assess their impact on IMDL, we consider two settings: context-aware, which relies only on surrounding content, and text-guided, which uses MS-COCO captions. While text guidance may fail with overly specific or misaligned captions, it generally improves fidelity and alignment. Quantitative evaluation with FID~\cite{FID} (Fig.~\ref{tab:fid}) shows that text-guided inpainting consistently outperforms the context-aware setting.

\begin{table}[t]
\centering
\small
\caption{COCO-Inpaint dataset composition. “\# w/ 48 variants” and “\# w/ 1 variant” indicate images with multiple and single inpainted variants, respectively. “Filtered” denotes invalid inpainted variants removed during quality filtering.}
\label{tab:dataset_composition}
\resizebox{\linewidth}{!}{
\begin{tabular}{lrrrrr}
\toprule
Subset   & Authentic & \# w/ 48 variants & \# w/ 1 variant & Filtered & Total Inpainted \\
\midrule
Training & 108,353   & 3,000              & 105,353         & 11,051 & 238,302         \\
Testing  & 4,551     & 125                & 4,426         & 491 & 9,935          \\
\bottomrule
\end{tabular}
}
\end{table}

\subsection{Scale, Diversity, and Quality}
To ensure broad coverage and evaluation under diverse conditions, we adopt a hybrid sampling strategy. From MS-COCO, we randomly select 3,000 images for training and 125 for validation, applying all 48 combinations of \textit{models} $\times$ \textit{masks} $\times$ \textit{prompts} ($6 \times 4 \times 2$) to each. The remaining images are processed using a single random combination to increase diversity while controlling dataset size. Original MS-COCO images are retained as authentic references. 
Dataset composition is shown in Tab.~\ref{tab:dataset_composition}.
This strategy yields a balanced and scalable benchmark with 238,302 inpainted and 108,353 authentic images (see Tab.~\ref{tab:dataset}). Tab.~\ref{tab:scale} further analyzes the influence of scale on model performance.
To reduce preprocessing bias, we apply a maximized CenterCrop to source images before inpainting. This mitigates the resizing artifacts caused by VAE, which require input dimensions divisible by 8 (or 16 for Flux.1).
Fig.\ref{fig:image_reward} compares ImageReward\cite{xu2023imagereward} distributions across datasets. COCO-Inpaint closely aligns with the original COCO, and outperforms TampCOCO, demonstrating higher visual quality and realism.
To ensure data quality, we adopt a two-stage post-processing strategy: automatic filtering with a binary classifier, followed by manual curation.

\begin{table}
\centering
\small
\caption{Cross-Model Generalization: ACC(\%), AUC(\%), and F1(\%) are reported. In-Distribution: Models are trained and evaluated on the same subset. Cross-Distribution: Models are trained on each training subset and evaluated on all testing subsets. Results are averaged across subsets.}
\vspace{-8pt}
\resizebox{\linewidth}{!}{
\begin{tblr}{
  row{2} = {c},
  cell{1}{1} = {r=2}{},
  cell{1}{2} = {c=2}{c},
  cell{1}{4} = {c=2}{c},
  cell{3}{2} = {c},
  cell{3}{3} = {c},
  cell{3}{4} = {c},
  cell{3}{5} = {c},
  cell{4}{2} = {c},
  cell{4}{3} = {c},
  cell{4}{4} = {c},
  cell{4}{5} = {c},
  cell{5}{2} = {c},
  cell{5}{3} = {c},
  cell{5}{4} = {c},
  cell{5}{5} = {c},
  cell{6}{2} = {c},
  cell{6}{3} = {c},
  cell{6}{4} = {c},
  cell{6}{5} = {c},
  cell{7}{2} = {c},
  cell{7}{3} = {c},
  cell{7}{4} = {c},
  cell{7}{5} = {c},
  vline{2-4} = {1}{dashed},
  vline{2,4} = {2}{dashed},
  vline{2,4} = {3-7}{dashed},
  hline{1,3,8} = {-}{},
}
Method       & In-Distribution         &                        & Cross-Distribution      &                                  \\
             & Pixel-Level             & Image-Level            & Pixel-Level             & Image-Level                      \\
IID          & 86.0/87.0/63.1          & 78.1/85.5/82.5         & 80.0/66.2/22.4          & 52.4/61.5/50.4                   \\
PSCC-Net     & 91.8/96.6/82.1          & \textbf{99.4/100/99.5} & 82.2/65.6/\textbf{24.6} & 52.0/70.8/31.9                   \\
ObjectFormer & 89.9/92.5/67.0          & 90.3/98.1/89.7         & \textbf{85.3}/66.4/19.0 & 61.2/70.8/37.4                   \\
IML-ViT      & 94.4/97.4/86.6          & 99.0/99.7/99.3         & 82.8/\textbf{66.9}/23.2 & 55.0/\textbf{75.4}/40.2          \\
SegFormer    & \textbf{95.9/98.4/89.7} & 90.2/100/93.3          & 83.2/63.2/23.5          & \textbf{62.1}/66.1/\textbf{62.6} 
\end{tblr}
}

\label{tab:model_avg}
\end{table}

\section{Experiments}
\label{sec:experiment}

\subsection{Settings}
\noindent\textbf{Baselines.}
We evaluate five representative IMDL methods: IID~\cite{IID_2021}, PSCC-Net~\cite{PSCC-Net_2022}, IML-ViT~\cite{IML-ViT_2023}, ObjectFormer~\cite{ObjectFormer_2022}, and SegFormer (MiT-B2)~\cite{SegFormer_2021}. All methods are tested under their official configurations, with Pixel-F1 early stopping (patience = 5).
For SegFormer, global average pooling is applied to the final-stage features, followed by a linear classification head. For IID and IML-ViT, which lack classification heads, we use the maximum value of the prediction map for binary classification. 

\noindent\textbf{Metrics.}
\label{sec:metrics}
We categorize evaluation into two levels: \textbf{Image-Level} and \textbf{Pixel-Level}.
Image-level metrics evaluate whether an input image has been manipulated (\textit{i.e.}, binary classification).
We use three common metrics: Accuracy (ACC), Area Under the Curve (AUC), and F1 score.
All image-level metrics are computed directly on the binary classification output.
Pixel-level evaluation focuses on the model’s ability to localize manipulated regions.
Metrics are computed pixel by pixel on ACC, AUC, and F1.
We adopt global pixel-level statistics for all pixel-level evaluation, avoiding instability and ensuring consistent, reliable evaluation.
Such inconsistencies have led to unfair evaluations in prior studies~\cite{PSCC-Net_2022, IML-ViT_2023}.

\begin{figure}[t]
    \centering
    \includegraphics[width=\linewidth]{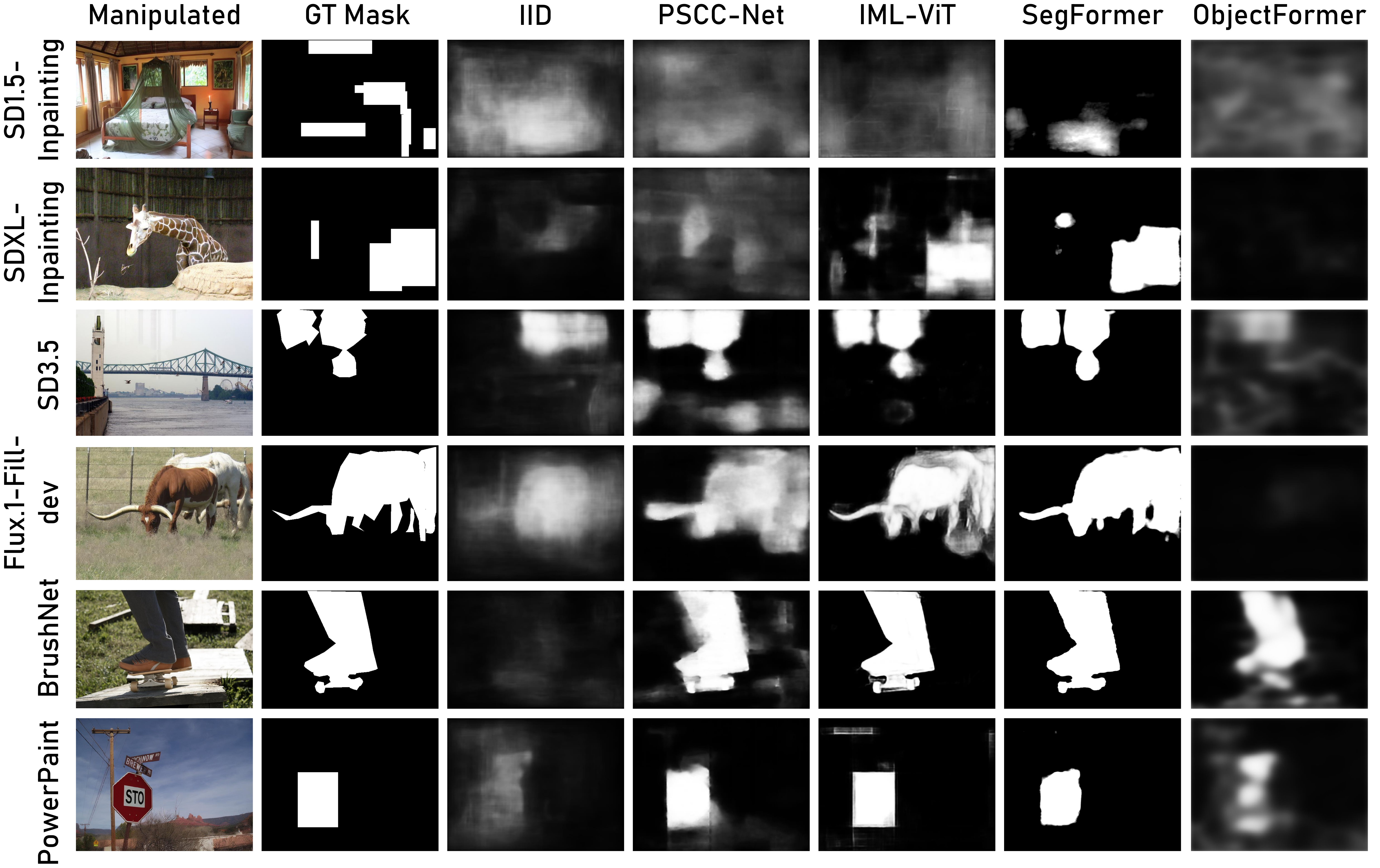}
    \vspace{-16pt}
    \caption{Qualitative IMDL results on COCO-Inpaint.}
    \label{fig:visual}
\end{figure}

\begin{figure}[t]
    \centering
    \includegraphics[width=\linewidth]{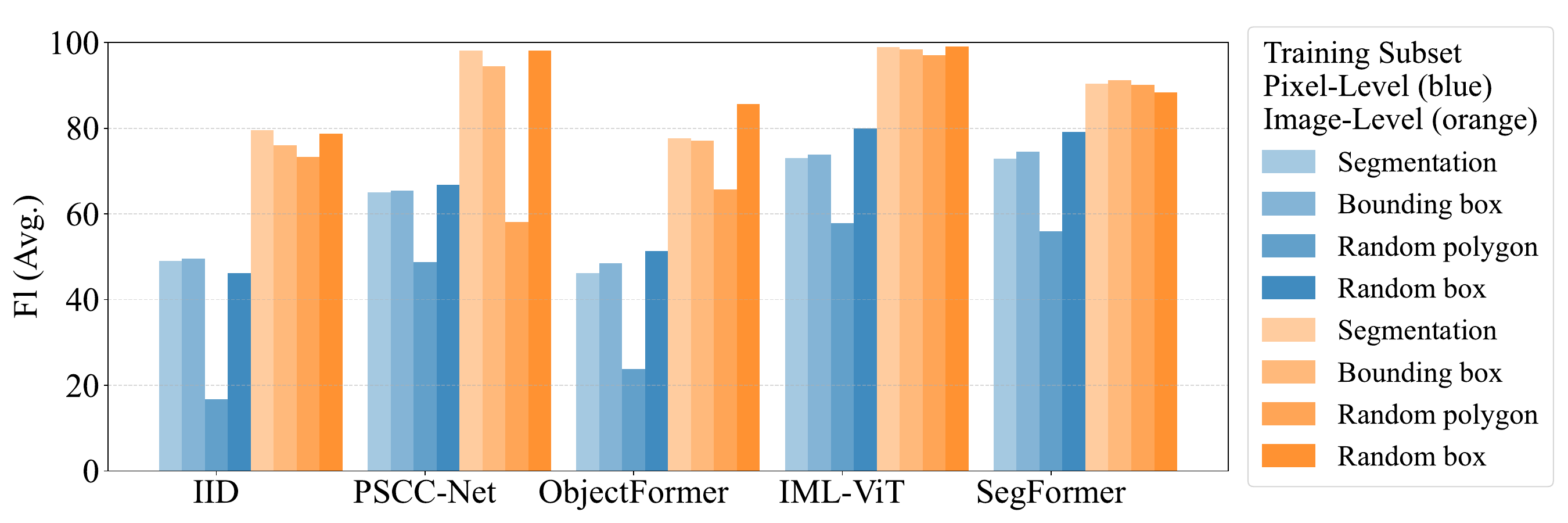}
    \vspace{-16pt}
    \caption{Cross-mask generalization averaged over mask types.}
    \label{fig:mask}
\end{figure}

\begin{table*}
  \centering
  \caption{Performance evaluation results. Models are trained on the full training set and evaluated on different testing subsets. ACC(\%), AUC(\%), and F1(\%) are reported. The best and second-best results are marked in \textbf{bold} and \underline{underline}, respectively.}
  \vspace{-8pt}
  \resizebox{\linewidth}{!}{
  \small
  \begin{tblr}{
    width = \linewidth,
    colspec = {Q[c, wd=0.08\linewidth] Q[c] Q[c] Q[c] Q[c] Q[c] Q[c] Q[c] Q[c] Q[c] Q[c] Q[c]},
    cells = {c},
    cell{1}{1} = {c=2,r=2}{0.121\linewidth},
    cell{1}{3} = {c=2}{0.16\linewidth},
    cell{1}{5} = {c=2}{0.16\linewidth},
    cell{1}{7} = {c=2}{0.16\linewidth},
    cell{1}{9} = {c=2}{0.162\linewidth},
    cell{1}{11} = {c=2}{0.16\linewidth},
    cell{3}{1} = {r=6}{},
    cell{9}{1} = {r=4}{},
    cell{13}{1} = {r=2}{},
    cell{15}{1} = {c=2}{0.121\linewidth},
    cell{16}{1} = {c=2}{0.121\linewidth},
    vline{3,5,7,9,11} = {1-16}{dashed},
    hline{1,17} = {-}{0.08em},
    hline{3,16} = {-}{},
    hline{9,13,15} = {-}{dotted},
  }
Testing Dataset      &                 & \makecell{IID~\cite{IID_2021}}           &                & \makecell{PSCC-Net~\cite{PSCC-Net_2022}}           &                                & \makecell{ObjectFormer~\cite{ObjectFormer_2022}}   &                & \makecell{IML-ViT~\cite{IML-ViT_2023}}                 &                                          & \makecell{SegFormer~\cite{SegFormer_2021}}              &                                        \\
                      &                 & Pixel-Level    & Image-Level    & Pixel-Level    & Image-Level                    & Pixel-Level    & Image-Level    & Pixel-Level             & Image-Level                              & Pixel-Level            & Image-Level                            \\
  Model               & SD1.5           & 83.9/83.8/56.0 & 72.7/83.0/77.4 & 84.5/88.7/62.7 & 96.9/99.8/97.7                 & 80.6/85.1/51.6 & 86.8/96.5/88.1 & \textbf{89.5/94.1/73.4} & \textbf{98.8/100/99.1}                   & \uline{84.9/92.8/69.5} & \uline{97.2/100/98.0}                  \\
                      & SDXL            & 86.2/86.7/63.3 & 75.5/84.8/79.9 & 91.9/97.3/81.9 & \uline{96.8}/99.9/\uline{97.6} & 84.0/87.5/55.8 & 85.3/93.7/86.6 & \textbf{97.2/99.6/93.2} & \textbf{99.9/100/99.9}                   & \uline{95.4/99.2/89.5} & 96.7/\uline{100}/97.6                  \\
                      & SD3.5           & 87.4/90.0/67.1 & 84.4/91.4/88.0 & 91.1/98.4/81.6 & \uline{97.6}/100/\uline{98.2}  & 84.2/86.7/56.1 & 81.3/90.9/82.1 & \textbf{98.1/99.8/95.3} & \textbf{99.9/100/99.9}                   & \uline{96.4/99.5/91.8} & 97.1/\uline{100}/97.8                  \\
                      & Flux.1-Fill-dev & 83.9/84.8/55.4 & 79.5/87.7/83.7 & 88.7/95.0/75.1 & \uline{97.1}/99.8/\uline{97.8} & 81.0/85.3/50.9 & 87.4/96.9/88.6 & \textbf{96.7/99.3/91.9} & \textbf{99.8/}\uline{99.9}\textbf{/99.9} & \uline{93.6/98.7/85.8} & 96.9/\textbf{100}/97.7                 \\
                      & BrushNet        & 91.9/94.4/78.3 & 83.0/91.0/86.6 & 97.8/99.7/94.7 & \uline{97.1}/99.7/\uline{97.9} & 90.5/94.7/71.5 & 84.1/92.7/85.1 & \textbf{99.5/99.7/98.8} & \textbf{100/100/100}                     & \uline{98.9/99.9/97.4} & 96.3/\uline{99.7}/97.3                 \\
                      & PowerPaint      & 85.0/85.9/55.9 & 73.7/84.5/78.1 & 91.6/96.2/80.0 & \uline{97.6}/100/\uline{98.2}  & 82.8/88.8/57.7 & 88.8/97.3/90.0 & \textbf{96.3/99.1/90.8} & \textbf{100/100/100}                     & \uline{94.3/98.8/87.0} & 96.9/\uline{100}/97.7                  \\
  Mask                & Segmentation    & 87.8/90.8/66.1 & 77.1/86.0/81.8 & 91.6/97.1/80.5 & \uline{97.0}/99.8/\uline{97.8} & 86.1/91.3/60.8 & 86.2/94.6/87.5 & \textbf{96.6/99.3/91.2} & \textbf{99.8/100/99.9}                   & \uline{94.5/98.9/87.1} & 96.8/\uline{99.9}/97.7                 \\
                      & Bounding box    & 88.3/93.9/77.8 & 82.6/90.5/86.7 & 93.4/98.4/88.8 & \uline{97.3}/99.9/\uline{98.0} & 87.8/94.0/73.3 & 87.0/96.2/87.6 & \textbf{97.4/99.7/95.4} & \textbf{99.8/100/99.9}                   & \uline{95.8/99.4/93.0} & 97.1/\uline{100}/97.9                  \\
                      & Random polygon  & 85.0/71.4/24.7 & 71.7/81.7/76.6 & 89.2/92.4/60.6 & 96.9/99.9/97.8                 & 81.5/79.5/32.2 & 86.5/94.6/87.4 & \textbf{95.6/98.0/80.6} & \textbf{99.5/}\uline{99.9}\textbf{/99.6} & \uline{92.7/97.5/73.9} & \uline{97.3}/\textbf{100}/\uline{98.1} \\
                      & Random box      & 83.6/85.4/58.7 & 81.0/89.2/85.1 & 89.2/95.6/77.9 & \uline{97.8}/99.9/\uline{98.4} & 83.1/88.3/57.1 & 87.4/95.6/87.8 & \textbf{95.3/98.9/89.5} & \textbf{99.8/100/99.9}                   & \uline{92.5/98.2/85.0} & 96.5/\uline{100}/97.4                  \\
  Text-guidance       & Prompt          & 84.8/86.0/58.6 & 76.9/86.0/81.9 & 90.5/96.3/79.1 & \uline{97.5}/99.9/\uline{98.2} & 83.2/87.0/55.4 & 85.0/94.4/86.7 & \textbf{96.1/99.1/90.5} & \textbf{99.8/100/99.8}                   & \uline{93.7/98.6/86.2} & 96.8/\uline{100}/97.7                  \\
                      & No-prompt       & 87.1/89.1/66.5 & 78.5/88.1/83.0 & 91.0/96.6/80.0 & 97.2/99.9/\uline{98.0}         & 83.7/88.4/58.7 & 86.6/95.5/88.0 & \textbf{96.3/99.2/91.0} & \textbf{99.7/100/99.8}                   & \uline{93.9/98.8/86.6} & \uline{97.3}/\uline{100}/98.1          \\
 \mbox{All Testing Dataset} &                 & 86.7/89.9/62.6 & 78.9/93.0/82.7 & 90.7/96.4/79.6 & \uline{97.4}/99.9/\uline{98.1} & 84.1/88.2/55.9 & 86.1/94.5/86.6 & \textbf{96.1/99.1/90.7} & \textbf{99.7/99.9/99.8}                  & \uline{93.7/98.7/86.4} & 97.1/\uline{99.9}/98.0                 
  \end{tblr}
  }
  \vspace{-8pt}
  \label{tab:all}
\end{table*}

\subsection{Results and Analysis}
\noindent \textbf{Full-Set Performance Evaluation.}
We first evaluate all baselines trained on the full training set. In addition to the full test set, we assess performance on specific subsets to better understand model behavior under different scenarios.
Quantitative results are presented in Tab.~\ref{tab:all}, with qualitative visualizations in Fig.~\ref{fig:visual}.
The results reveal that CNNs exhibit limited performance on inpainting-based manipulations.
In contrast, ViT-based architectures (IML-ViT and SegFormer) demonstrate superior robustness by modeling global dependencies through self-attention, enabling the detection of subtle and spatially dispersed manipulations.
This advantage is particularly evident when manipulated regions are small (e.g., mask ratio $\textless$ 0.2), where CNN-based methods such as PSCC often predict near-zero masks, highlighting the importance of global context modeling when local artifacts are sparse.
ObjectFormer, due to its explicit learning of mid-level object representations, struggles with complex and diverse mask shapes, resulting in suboptimal performance.
For different mask types, object-based manipulations (segmentation and bounding box) are easier to localize than random masks.
\textit{E.g.}, the F1 scores of ObjectFormer on random inpainted images (0.322 and 0.571) are notably lower than those on object-based ones (0.608 and 0.733), corroborating this observation.
This is likely because random masks disrupt object--semantic alignment, hindering the use of object-level semantics for localization.
Finally, text-guided inpainting improves realism but slightly increases localization difficulty across all metrics.

\noindent \textbf{Generalization Evaluation Across Models.}
Given the rapid evolution of inpainting architectures, generalization to unseen models is crucial for IMDL.
To evaluate this, COCO-Inpaint is divided into six subsets by inpainting model, each with separate training and test splits. 
All IMDL methods are trained and cross-evaluated across these splits.
Tab.~\ref{tab:model_avg} reports the average performance under in-distribution and cross-model settings. 
ViT-based models consistently outperform CNNs in both classification and localization. However, localization performance drops sharply across models, with Pixel-F1 reaching only 24.6\% due to style discrepancies and distribution shifts. These results highlight poor cross-model generalization, which we attribute to overfitting to generator-specific artifacts.
\begin{figure}[t]
    \centering
    \includegraphics[width=\linewidth]{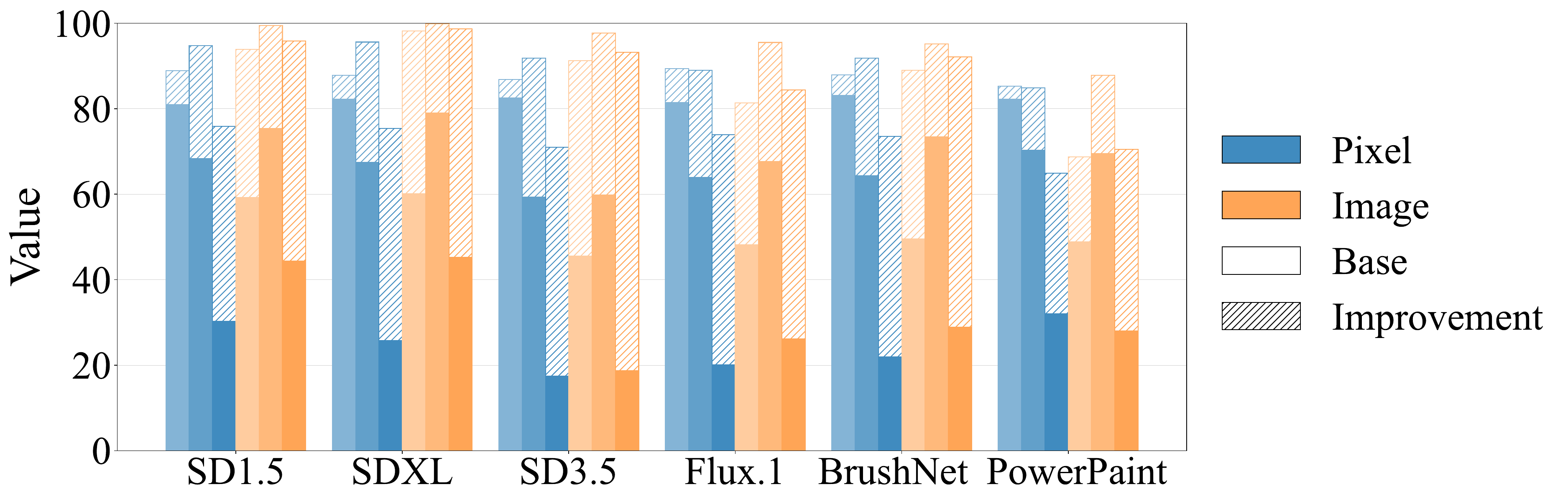}
    \vspace{-16pt}
    \caption{The results of PSCC-Net trained on the remaining five data subsets and tested on the current subset. The six bars represent Pixel ACC/AUC/F1, Image ACC/AUC/F1, respectively. Improvement indicates the relative gain when trained on a single subset.}
    \label{fig:pscc_train5_test1}
\end{figure} 
To address this limitation, we advocate training on multiple inpainting models to encourage learning of common manipulation artifacts. As shown in Fig.~\ref{fig:pscc_train5_test1}, PSCC-Net trained on five subsets (\textit{i.e.}, SDXL, SD3.5, Flux, BrushNet, and PowerPaint) and tested on the remaining one (\textit{i.e.}, SD1.5) generalizes significantly better to unseen distributions.
This cross-model training protocol offers a practical strategy for improving robustness against future inpainting techniques. \looseness= -1

\begin{table}[t]
\centering
\small
\caption{SegFormer pretrained on COCO-Inpaint evaluated on the CNN-syn. subset of HiFi-IFDL. OSN-det. and CNN-det., pretrained on their own datasets, are the strongest baselines within HiFi-IFDL.}
\setlength{\tabcolsep}{6pt} 
\begin{subtable}{0.48\linewidth}
\centering
\begin{tabular}{lcc}
\hline
Model     & AUC  & F1   \\
\hline
OSN-det.  & 51.4 & 38.8 \\
SegFormer & 84.5 & 34.6 \\
\hline
\end{tabular}
\caption{Pixel-level}
\label{tab:hifi-ifdl_pixel}
\end{subtable}
\hfill
\begin{subtable}{0.48\linewidth}
\centering
\begin{tabular}{lcc}
\hline
Model     & AUC  & F1   \\
\hline
CNN-det.  & 76.5 & 60.5 \\
SegFormer & 99.3 & 95.1 \\
\hline
\end{tabular}
\caption{Image-level}
\label{tab:hifi-ifdl_image}
\end{subtable}

\label{tab:hifi-ifdl}
\end{table}

\noindent \textbf{Cross-Domain Generalization Evaluation.} 
To assess cross-domain robustness, we evaluate SegFormer pretrained on COCO-Inpaint using the partially manipulated subsets (CNN-syn.) of HiFi-IFDL~\cite{HiFi-IFDL_2023}, where CNN-syn. specifically targets facial manipulations. In this setting, \textbf{SegFormer achieves 84.5\% Pixel-AUC for segmentation and 99.3\% Image-AUC for classification, demonstrating strong generalization}. 
SegFormer substantially improves Pixel-AUC and Image-level AUC/F1 over the HiFi-IFDL baselines, although its Pixel-F1 remains slightly lower than OSN-det.

\noindent \textbf{Generalization Evaluation Across Masks.}
Inpainting models are often trained using different mask strategies~\cite{BrushNet_2024}, often relying on segmentation datasets for mask extraction.
LaMa~\cite{LaMa_2022} argues that mask type greatly impacts inpainting model performance.
To assess robustness, we conduct cross evaluation across mask types (Fig.~\ref{fig:mask}). 
Models trained with \textbf{random box masks generalize the best}, likely because random box masks decouple manipulated regions from object semantics, unlike segmentation or bounding boxes.
Irregular polygons are most challenging due to structural ambiguity.

\begin{table}[t]
\centering
\small
\caption{Dataset scale effect results. Models trained on different fractions and tested on the full set. Pixel-F1/Image-F1 are reported.}
\vspace{-8pt}
\setlength{\tabcolsep}{3pt} 
\resizebox{\linewidth}{!}{
\begin{tabular}{cccccc} 
\hline
Scale & IID                         & PSCC                       & ObjectFormer                & IML-ViT                    & SegFormer                    \\ 
\hline
1               & \textbf{62.6}/\textbf{82.7} & \textbf{79.6}/98.1         & \uline{55.9}/\uline{86.6}   & \textbf{90.7}/\uline{99.8} & \textbf{86.4}/\textbf{98.0}  \\
1 / 2           & \uline{59.6}/\uline{80.1}   & \uline{76.4}/\textbf{99.3} & \textbf{56.1}/\textbf{86.8} & \uline{89.2}/\textbf{99.9} & \uline{85.7}/\uline{94.9}    \\
1 / 4           & 50.6/70.8                   & 71.8/\uline{99.1}          & 49.9/79.1                   & 87.4/99.8                  & 81.2/86.8                    \\
1 / 8           & 46.9/67.9                   & 67.8/95.6                  & 47.1/76.7                   & 83.7/99.3                  & 80.7/92.5                    \\
1 / 16          & 38.6/69.2                   & 61.0/80.9                  & 42.2/76.1                   & 39.7/81.3                  & 77.4/91.4                    \\
\hline
\end{tabular}
}
\label{tab:scale}
\end{table}

\begin{figure}[t]
    \centering
    \includegraphics[width=\linewidth]{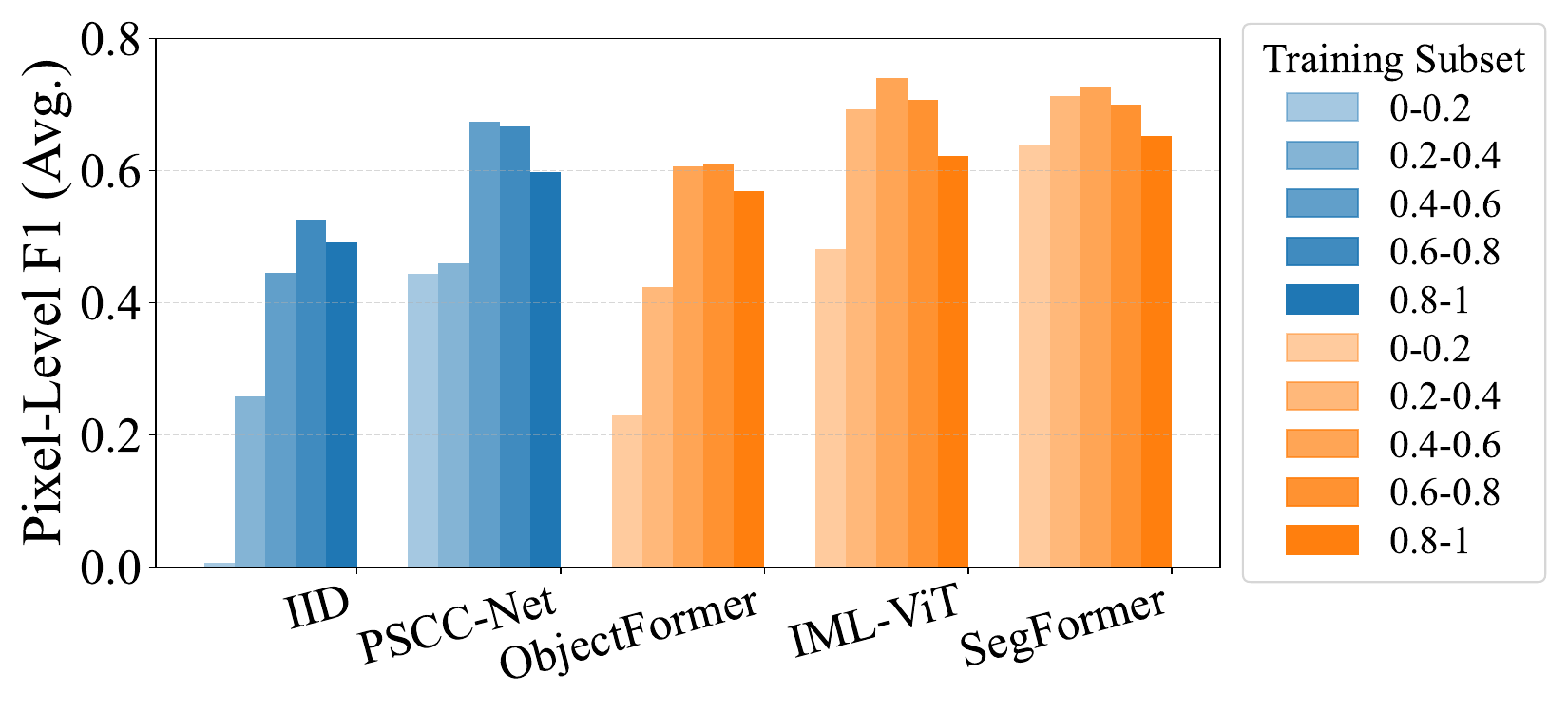}
    \vspace{-16pt}
    \caption{Cross-Mask-Ratio Generalization: Average pixel-level F1 scores of IMDL models under different mask ratios. Colors distinguish between architectures (CNNs vs. ViTs)} 
    \label{fig:PixelLevel_F1}
\end{figure}

\noindent \textbf{Generalization Evaluation Across Mask Ratio.}
Mask area has a strong impact on inpainting performance~\cite{LaMa_2022}.
To study its effect on IMDL, we conduct a cross-evaluation across mask ratio (Fig.~\ref{fig:PixelLevel_F1}).
Each subset randomly samples 18,000 training images to control scale, while testing on the full-set.
\textbf{Most IMDL models achieve optimal performance at mask ratios of 0.4–0.6.}
This indicates that IMDL performs best when authentic and manipulated regions are balanced.

\noindent \textbf{Generalization Evaluation Across Prompts.}
\begin{figure}[t]
    \centering
    \includegraphics[width=\linewidth]{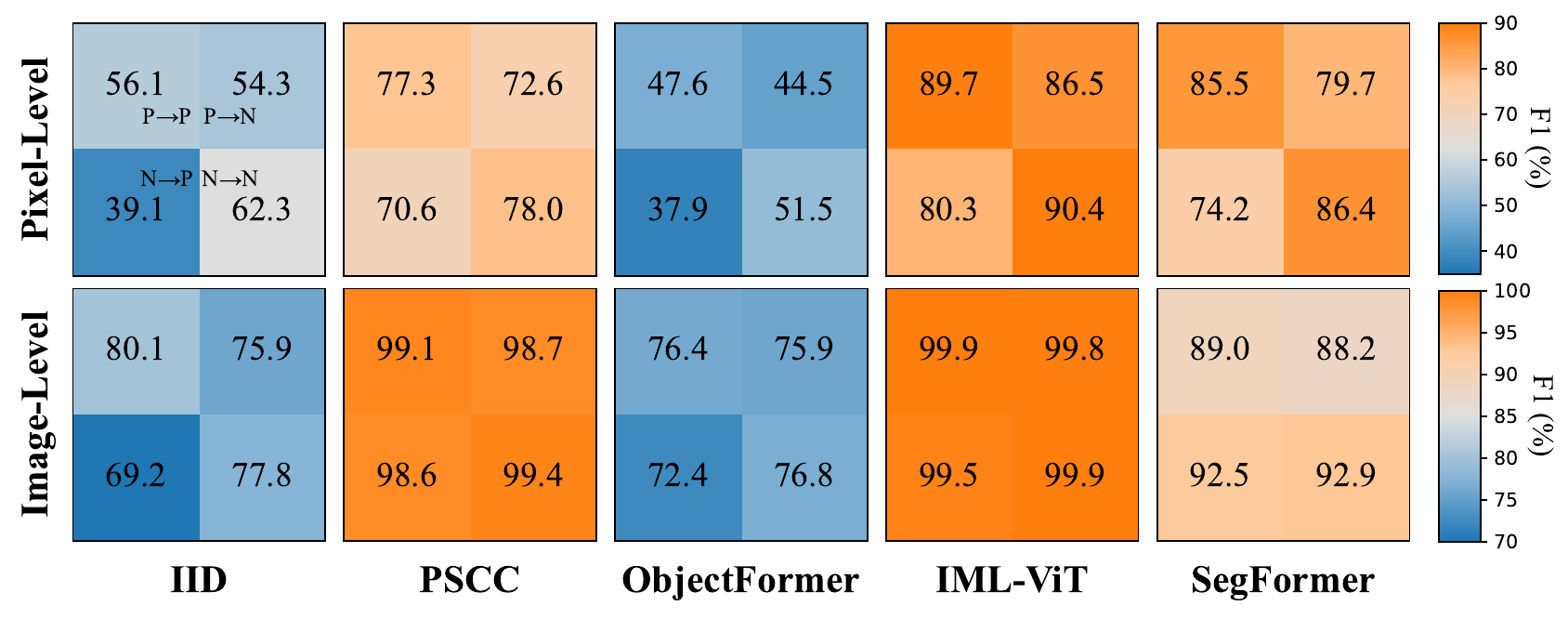}
    \vspace{-16pt}
    \caption{Cross-Text-Guidance Generalization. F1 scores. $P \rightarrow N$ denotes training with prompts and testing without prompts.}
    \label{fig:cross_prompt}
\end{figure}
Modern inpainting models can operate with or without text prompts~\cite{Flux_2024,SDXL_2023,SD_2022}.
To analyze its impact on IMDL, we conduct cross-validation on prompt and no-prompt subsets (Fig.~\ref{fig:cross_prompt}), showing that no-prompt training achieves higher in-distribution scores, while prompt training improves robustness to unseen distributions by supplying richer semantic context.

\noindent \textbf{Analysis of the Benchmark Scale.}
\label{subsec:dataset_scale}
We analyze scaling effects by training on random subsets of COCO-Inpaint. Performance increases with dataset size, while the marginal gains diminish beyond half-scale (Tab.~\ref{tab:scale}).

\section{Conclusion}
\label{sec:conclusion}
In this paper, we propose COCO-Inpaint, a benchmark for Inpainting Image Manipulation Detection and Localization (IMDL).
COCO-Inpaint utilizes 6 inpainting models (SD1.5-Inpainting, SDXL-Inpainting, SD3.5, Flux.1-Fill-dev, BrushNet, and PowerPaint) to generate manipulated images by 4 mask types (Segmentation, Bounding Box, Random Polygon, Random box) with two prompt guidance types.
Based on the hierarchical structure of COCO-Inpaint, multi-level performance and generalization evaluation experiments are conducted, providing a comprehensive and in-depth analysis of the performance of existing IMDL methods in handling inpainting-based manipulated images.

\section{Acknowledgments}
This work was supported in part by the National Natural Science Foundation of China (Grant Nos. 62302295, 62595733, and 62561160155), the Shanghai Municipal Science and Technology Major Project (Grant No. 2021SHZDZX0102). This work was also supported by Ant Group.

\bibliographystyle{IEEEbib}
\bibliography{strings,refs}

\end{document}